\documentclass[10pt,twocolumn,letterpaper]{article}

\usepackage{times}
\usepackage{epsfig}
\usepackage{graphicx}
\usepackage{amsmath}
\usepackage{amssymb}

\usepackage{amsfonts}
\usepackage{amsmath}
\usepackage{graphicx}
\usepackage{caption}
\usepackage{subcaption}
\usepackage{algorithm}
\usepackage{algpseudocode}
\usepackage{multirow}
\usepackage{array}
\usepackage{float}
\usepackage[OT1]{fontenc}

\usepackage[pagebackref=true,breaklinks=true,letterpaper=true,colorlinks,bookmarks=false]{hyperref}

\def\cvprPaperID{2362} 

\begin{document}

\title{Show, Attend and Translate: Unpaired Multi-Domain \\Image-to-Image Translation with Visual Attention}

\author{
Honglun Zhang$^{1}$, Wenqing Chen$^{1}$, Jidong Tian$^{1}$, Yongkun Wang$^{2}$, Yaohui Jin$^{1}$ 
\\ 
$^{1}$State Key Lab of Advanced Optical Communication System and Network\\
MoE Key Lab of Artificial Intelligence, AI Institute, Shanghai Jiao Tong University \\
$^{2}$Network and Information Center, Shanghai Jiao Tong University \\ 
\{jinyh\}@sjtu.edu.cn
}

\maketitle

\begin{abstract}
Recently unpaired multi-domain image-to-image translation has attracted great interests and obtained remarkable progress, where a label vector is utilized to indicate multi-domain information. In this paper, we propose \textbf{SAT}~(Show, Attend and Translate), an unified and explainable generative adversarial network equipped with visual attention that can perform unpaired image-to-image translation for multiple domains. By introducing an action vector, we treat the original translation tasks as problems of arithmetic addition and subtraction. Visual attention is applied to guarantee that only the regions relevant to the target domains are translated. Extensive experiments on a facial attribute dataset demonstrate the superiority of our approach and we utilize the residual images to better explain what SAT attends when translating images.
\end{abstract}

\section{Introduction}

Recently image-to-image translation has attracted great interests and obtained remarkable progress with the prosperities of generative adversarial networks~(GANs)~\cite{DBLP:conf/nips/GoodfellowPMXWOCB14}. It aims to change a certain aspect of a given image in a desired manner and covers a wide variety of applications, ranging from changing face attributes like hair color and gender~\cite{DBLP:journals/corr/abs-1711-09020}, reconstructing street scenes from semantic label maps~\cite{DBLP:conf/cvpr/IsolaZZE17}, to transforming realistic photos into art works~\cite{DBLP:journals/corr/ZhuPIE17}.

A \textbf{domain} refers to a group of images that share some latent semantic features in common, which are denoted as domain \textbf{labels}. The values of different domain labels can be either \textit{binary}, like male and female for gender, or \textit{categorical} such as black, blonde and brown for hair color. Based on the above definition, the scenarios and applications of image-to-image translation for two domains are extremely various, interesting and creative~\cite{DBLP:journals/corr/ZhuPIE17,DBLP:conf/cvpr/IsolaZZE17,DBLP:conf/icml/KimCKLK17,DBLP:conf/nips/LiuBK17}.

A more complicate and useful challenge, however, is multi-domain image-to-image translation, which is supposed to transform a given image to several target domains of high qualities. There are a few benchmark image datasets available with more than two labels. For example, the CelebA~\cite{liu2015faceattributes} dataset contains about 200K celebrity face images, each annotated with 40 binary labels describing facial attributes like hair color, gender and age. Inspired by \cite{DBLP:journals/corr/ZhuPIE17}, \cite{DBLP:journals/corr/abs-1711-09020} utilizes a \textbf{label vector} to convey the information of multiple domains, and applies \textbf{cycle consistent loss} to guarantee preservations of domain-unrelated contents. However, there are not many other literature dedicated to this topic, still leaving a lot of room for improvements.

In this paper, we propose the model \textbf{SAT}~(Show, Attend and Translate), an unified and explainable generative adversarial network to achieve unpaired multi-domain image-to-image translation with a single generator. We explore different fashions of combining the given images with the multi-domain information, either in the \textbf{raw} or \textbf{latent} phase, and compare their influences on the translation results. Based on the label vector, we propose the \textbf{action vector}, a more intuitive and understandable representation that converts the original translation tasks as problems of arithmetic addition and subtraction. We utilize the effective \textbf{visual attention}~\cite{DBLP:conf/icml/XuBKCCSZB15} to capture the correlations between the given images and the target domains, so that the domain-unrelated regions are preserved. 

We conduct extensive experiments and the results demonstrate the superiority of our approach. The residual images are utilized to better visualize what SAT attends when translating images. 

Our contributions are summarized in three-folds:

\begin{itemize}
    \item We propose SAT, an unified and explainable generative adversarial network for unpaired multi-domain image-to-image translation. 
    \item SAT utilizes the action vector to convey the information of target domains and the visual attention to determine which regions to focus when translating images.
    \item Both qualitative and quantitative experiments on a facial attribute dataset demonstrate the effectiveness of our approach.
\end{itemize}

\begin{figure*}[!ht]
\centering
\includegraphics[width=0.9\textwidth]{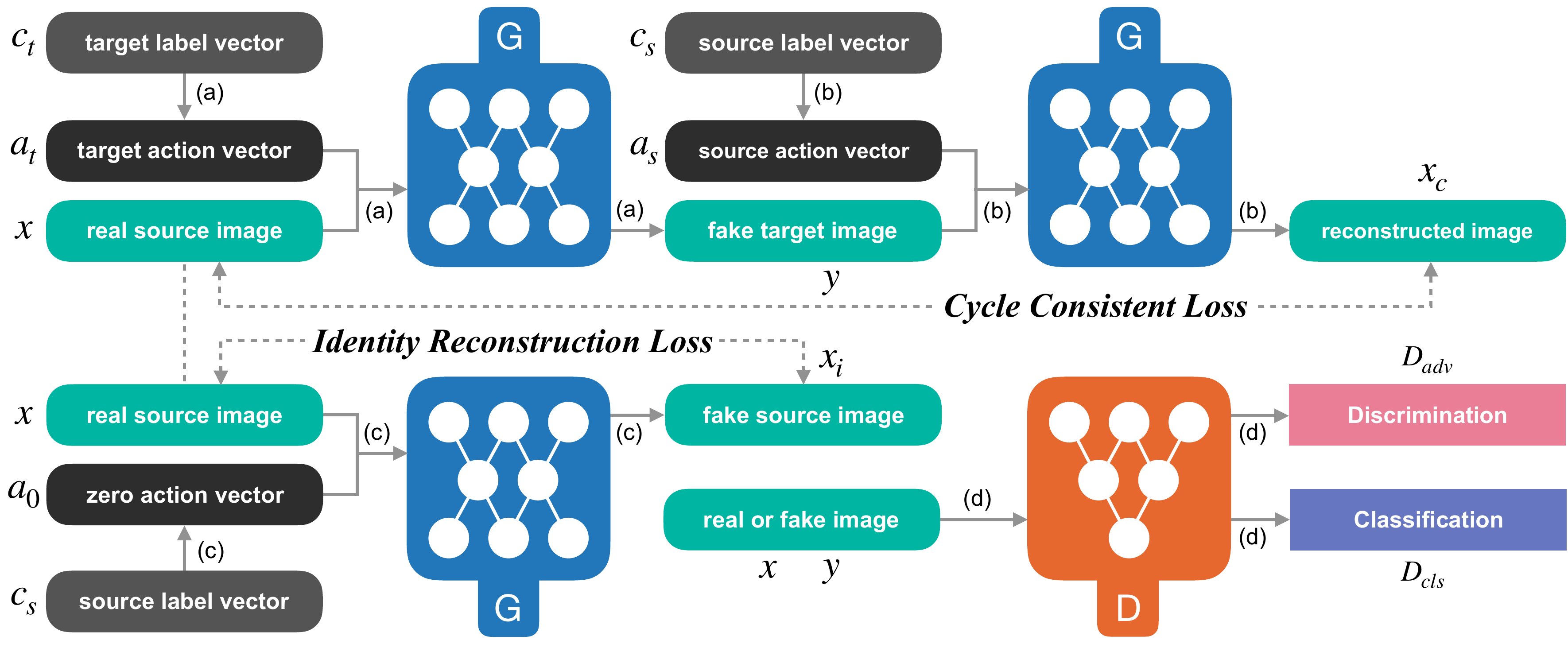}
\caption{Overall architecture of Show, Attend and Translate, which consists of a generator $G$ and a discriminator $D$. \textbf{(a)} $G$ takes as input both the real source image $x$ and the target action vector $a_t$ to synthesize the fake target image $y$. \textbf{(b)} $G$ tries to obtain the reconstructed image $x_c$ based on the fake target image $y$ and the source action vector $a_s$, where the \textit{Cycle Consistent Loss} is imposed on $x$ and $x_c$. \textbf{(c)} $G$ accepts the real source image $x$ and a zero action vector $a_0$ to produce the fake source image $x_i$, which should be exactly the same with $x$ constrained by the \textit{Identity Reconstruction Loss}. \textbf{(d)} $D$ learns to distinguish the fake images from the real ones and infer the most appropriate labels for classification.}\label{architecture1}
\end{figure*}

\section{Related Work}

\textbf{Generative Adversarial Nets}. Generative Adversarial Nets~(GANs)~\cite{DBLP:conf/nips/GoodfellowPMXWOCB14} are a powerful method for training generative models of complicate data and have been proven effective in a wide variety of applications, including image generation~\cite{DBLP:journals/corr/RadfordMC15,DBLP:conf/iccv/ZhangXL17,DBLP:conf/nips/GulrajaniAADC17}, image-to-image translation~\cite{DBLP:conf/cvpr/IsolaZZE17,DBLP:journals/corr/ZhuPIE17,DBLP:journals/corr/abs-1711-09020}, image super-resolution~\cite{DBLP:conf/cvpr/LedigTHCCAATTWS17} and so on. Typically a GAN model consists of a \textit{Generator}~($G$) and a \textit{Discriminator}~($D$) playing a two-player game, where $G$ tries to synthesize fake samples from random noises following a prior distribution, while $D$ learns to distinguish those from real ones. The two roles combat with each other and finally reach a \textit{Nash Equilibrium}, where the generator is able to produce indistinguishable fake samples of high qualities.

\textbf{Auxiliary Classifier GANs}. Several work are devoted to controlling certain details of the generated images by introducing additional supervisions, which can be a multi-hot label vector indicating the existences of some target attributes~\cite{DBLP:journals/corr/MirzaO14,DBLP:conf/icml/OdenaOS17}, or a textual sentence describing the desired content to generate~\cite{DBLP:conf/iccv/ZhangXL17,DBLP:journals/corr/abs-1710-10916,DBLP:journals/corr/abs-1711-10485}. The Auxiliary Classifier GANs~(ACGAN)~\cite{DBLP:conf/icml/OdenaOS17} belongs to the former group and the label vector conveys semantic implications such as gender and hair color in a facial synthesis task. $D$ is also enhanced with an auxiliary classifier that learns to infer the most appropriate label for any real or fake sample. Based on the label vector, we further propose the action vector, which is more intuitive and explainable for image-to-image translation.

\textbf{Image-to-Image Translation}. There is a large body of literature dedicated to image-to-image translation with impressive progress. For example, pix2pix~\cite{DBLP:conf/cvpr/IsolaZZE17} proposes an unified architecture for paired image-to-image translation based on cGAN~\cite{DBLP:journals/corr/MirzaO14} and a L1 reconstruction loss. To alleviate the costs for obtaining paired data, the problem of unpaired image-to-image translation has also been widely exploited~\cite{DBLP:journals/corr/ZhuPIE17,DBLP:conf/icml/KimCKLK17,DBLP:conf/nips/LiuBK17}, which mainly focuses on translating images between two domains. For the more challenge task of multi-domain image-to-image translation, StarGAN~\cite{DBLP:journals/corr/abs-1711-09020} combines the ideas of \cite{DBLP:journals/corr/ZhuPIE17} with \cite{DBLP:conf/icml/OdenaOS17} and can robustly translate a given image to multiple target domains with only a single generator. In this paper, we dive deeper into this issue and integrate several novel improvements.

\textbf{Visual Attention}. Attention based models have demonstrated significant performances in a wide range of applications, including neural machine translation~\cite{DBLP:conf/nips/VaswaniSPUJGKP17}, image captioning~\cite{DBLP:conf/icml/XuBKCCSZB15}, image generation~\cite{DBLP:journals/corr/abs-1805-08318} and so on. \cite{DBLP:conf/eccv/ZhangKSC18} utilizes the visual attention to localize the domain-related regions for facial attribute editing, but can only handle a single attribute by translating between two domains. In this paper, we validate the use of attention to solve the more generalized problem of translating images for multiple domains.

\begin{figure*}[!ht]
\centering
\includegraphics[width=0.9\textwidth]{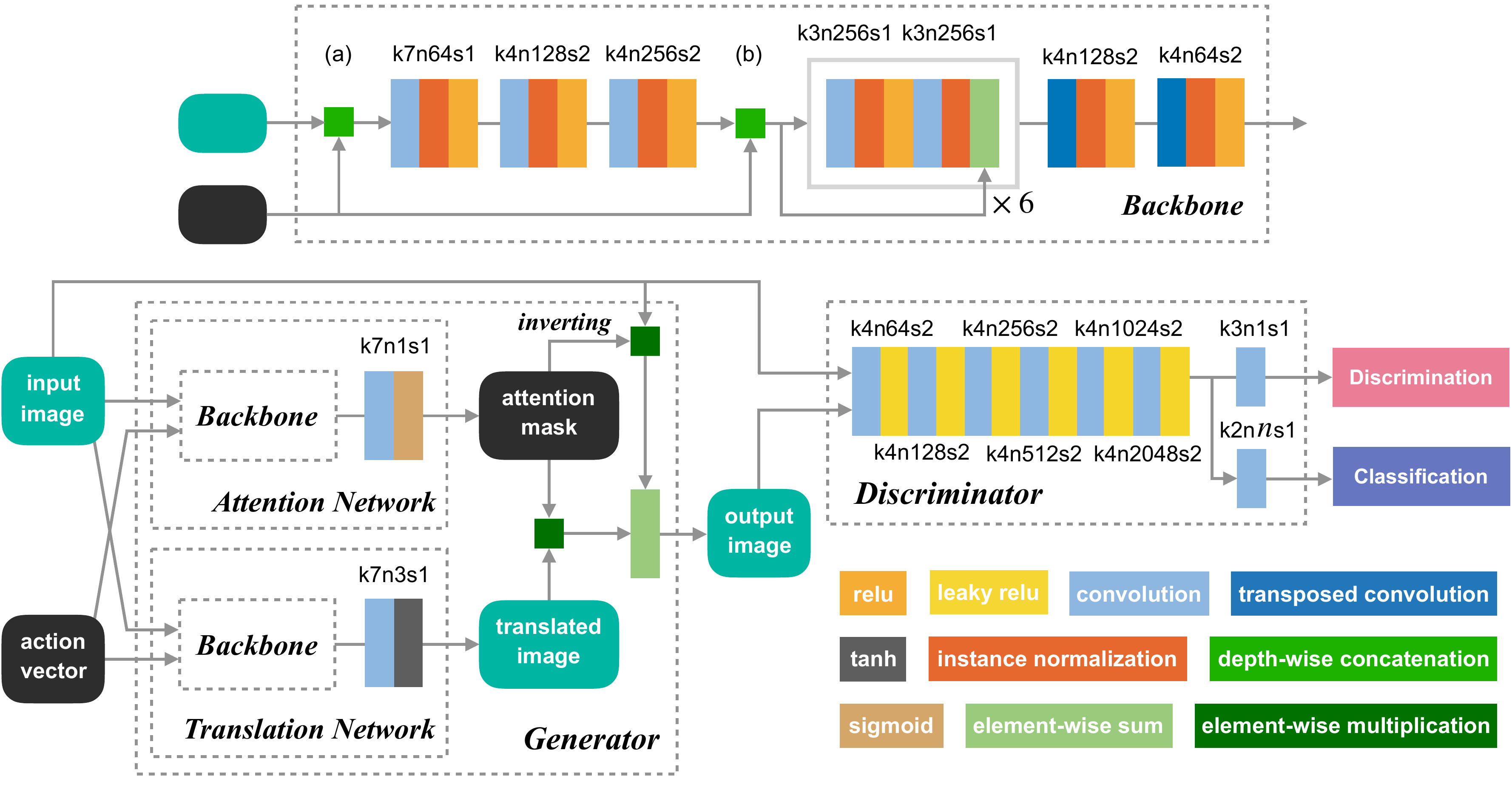}
\caption{Detailed network architectures of $G$ and $D$ in SAT. $G$ consists of two sub-modules, the Translation Network and the Attention Network. $D$ produces probability distributions for both adversarial discrimination and label classification. (a) and (b) denote the raw and latent strategies of combination in the backbone respectively.}\label{architecture2}
\end{figure*}

\section{Methodology}

In this section, we discuss our approach \textbf{SAT}~(Show, Attend and Translate) for unpaired multi-domain image-to-image translation. The overall and detailed architectures of SAT are illustrated in Fig.\ref{architecture1} and Fig.\ref{architecture2} respectively.

\subsection{Multi-Domain Image-to-Image Translation}

Multi-domain image-to-image translation accepts an input image $x$ and produces an output image $y$ for multiple target domains. We denote the number of involved domains by $n$ and utilize a multi-hot \textbf{label vector} $c_t\in\{0,1\}^n$ to convey the target domain information, where $1$ means the existence of a certain domain while $0$ acts oppositely.

\subsection{Raw or Latent}

Generally $G$ consists of three parts, an \textit{encoder} to downsample $x$ to a latent representation $\tilde{x}$ of lower resolution, several residual blocks~\cite{DBLP:conf/cvpr/HeZRS16} for nonlinear transformations, and a \textit{decoder} for upsampling and producing $y$ of the original resolution. So there are two different strategies of combining the input image with the target domain information.

\begin{itemize}
    \item \textbf{Raw}. Combine $x,c_t$ before the encoder by spatially replicating $c_t$, which is then concatenated with $x$.
    \item \textbf{Latent}. Combine $\tilde{x},c_t$ after the encoder and before the residual blocks in a similar way.
\end{itemize}

\cite{DBLP:journals/corr/abs-1711-09020} adopts the former manner, but we suspect that it may be more intuitive to introduce $c_t$ in the latent phase, as $c_t$ contains the semantic implications about the target domains, so it should be less appropriate to directly combine $c_t$ with the raw pixel values of $x$. We will further investigate this issue in the Experiments section.

\subsection{Action Vector}

In \cite{DBLP:conf/icml/OdenaOS17,DBLP:journals/corr/abs-1711-09020}, the label vector $c_t$ is utilized to contain the target domain information. Based on $c_t$, we propose the \textbf{action vector} for guiding $G$ to translate images corresponding to the target domains. The label vector of the source domains is denoted by $c_s$, so we need to translate an input image $x$ with $c_s$ to an output image $y$ with $c_t$. The target action vector $a_t\in\{-1,0,1\}^n$ is defined as follows. 
\begin{equation}\tag{$1$}\label{eq:1}
a_t=c_t-c_s
\end{equation}

The motivation of action vector is straightforward and meaningful. $c_t$ tells the model how the generated images should look like, while $a_t$ describes what should be done and changed to generate the desired outputs. Given that $c_s$ and $c_t$ are both multi-hot vectors, the values of $a_t$ should be $-1$, $0$ or $1$, which mean \textit{removing}, \textit{preserving} or \textit{adding} the related content of a target domain respectively. In this way, the original task of multi-domain image-to-image translation can be understood as a problem of arithmetic addition and subtraction, $G(x,a_t)\rightarrow y$, enabling the model to focus more on what should be changed and translated for the target domains.

Based on $y$ and $c_t$, we can also conduct the translation reversely conditioned on $c_s$ as Fig.\ref{architecture1} shows and obtain the reconstructed image, $G(y,a_s)\rightarrow x_c$, where the source action vector $a_s$ is calculated as follows.
\begin{equation}\tag{$2$}\label{eq:2}
a_s=c_s-c_t
\end{equation}

It is noticeable that $a_s=-a_t$, which well coincides with the definition of reverse translation. We use $a_0\in \{0\}^n$ to denote the \textbf{zero action vector}, which means translating nothing and preserving the content for the original domains.

\subsection{Visual Attention}

In order to leverage the effective visual attention, we modify the generator and now $G$ consists of two sub-modules, the Translation Network~(\textbf{TN}) and the Attention Network~(\textbf{AN}) as Fig.\ref{architecture2} shows. TN achieves the translation task and generates ${y}'$ from $x$ conditioned on $a_t$.
\begin{equation}\tag{$3$}\label{eq:3}
{y}'=T(x,a_t)
\end{equation}

However, TN may change the domain-unrelated contents and thus produce unsatisfactory results, which should be avoided for multi-domain image-to-image translation. To solve this problem, AN accepts $x,a_t$ and generate an attention mask $M$ with the same resolution as $x$.
\begin{equation}\tag{$4$}\label{eq:4}
M=A(x,a_t)
\end{equation}

Ideally, the values of $M$ should be either $0$ or $1$, which indicates that the corresponding pixel of the input image $x$ is irrelevant or relevant to the target domains. Based on the attention mask, we can obtain the refined translation result $y$ by extracting only the domain-related content from ${y}'$ and copying the rest from $x$.
\begin{equation}\tag{$5$}\label{eq:5}
y=M\cdot {y}' + (1-M)\cdot x
\end{equation}
where $\cdot$ denotes element-wise multiplication and $1-M$ means inverting the attention mask to get the domain-unrelated regions. We combine Eq.(\ref{eq:3})-(\ref{eq:5}) and reformulate $G$ as follows.
\begin{equation}\tag{$6$}\label{eq:6}
y=G(x,a_t)=(1-A(x,a_t))\cdot x+A(x,a_t)\cdot T(x,a_t)
\end{equation}

In this way, we decompose the original task of multi-domain image-to-image translation into two sub-tasks, determining which regions to focus attention on and learning how to generate realistic images conforming to the target domains, which guarantee that the domain-unrelated contents are preserved when translating images. The detailed network architectures of $G,D$ are illustrated in Fig.\ref{architecture2} and will be further discussed in the Implementation section.

\subsection{Loss Functions}

\noindent\textbf{Adversarial Loss}. In order to distinguish fake samples from real ones, $D$ learns to minimize the following adversarial loss~\cite{DBLP:conf/nips/GoodfellowPMXWOCB14}.
\begin{equation}\tag{$7$}\label{eq:7}
\begin{split}
\mathcal{L}_{adv}^D=&-\mathbb{E}_x[\log D_{adv}(x)]\\&-\mathbb{E}_{x,a_t}[\log(1-D_{adv}(G(x,a_t)))]
\end{split}
\end{equation}
while $G$ tries to synthesize fake samples to fool $D$ so the adversarial loss of $G$ acts oppositely.
\begin{equation}\tag{$8$}\label{eq:8}
\mathcal{L}_{adv}^G=\mathbb{E}_{x,a_t}[\log(1-D_{adv}(G(x,a_t)))]
\end{equation}

By playing such a two-player game, $D$ obtains stronger capability of discrimination and $G$ is able to generate realistic samples of high qualities.\vspace{2mm}

\noindent\textbf{Classification Loss}. In order to translate the input image $x$ to the desired output $y$ conforming to the target domains $c_t$, $D$ should possess the capability of classifying images to their correct labels, and $G$ should be able to generate images corresponding to $c_t$ with supervisions from $D$. 

We utilize the annotations between the source images $x$ and the source label vectors $c_s$ to train the auxiliary classifier in $D$. By minimizing the following classification loss, where $c_s$ are the ground truths and $D_{cls}(c_s|x)$ are the predicted probability distributions, $D$ learns to infer the most appropriate labels with confidence for any given images.
\begin{equation}\tag{$9$}\label{eq:9}
\mathcal{L}_{cls}^D=\mathbb{E}_{x,c_s}[-\log D_{cls}(c_s|x))]
\end{equation}

At the meanwhile, we also impose the classification loss on $G$ to guarantee that the generated images $G(x,a_t)$ are not only realistic but also classified to the target domains $c_t$ supervised by $D$.
\begin{equation}\tag{$10$}\label{eq:10}
\mathcal{L}_{cls}^G=\mathbb{E}_{x,a_t,c_t}[-\log D_{cls}(c_t|G(x,a_t)))]
\end{equation}

\noindent\textbf{Cycle Consistent Loss}. It is intuitive that multi-domain image-to-image translation should only change the domain-related contents while keeping other details preserved, which cannot be solely satisfied by Eq.(\ref{eq:8}) and (\ref{eq:10}). After translation and reverse translation, the reconstructed image $x_c$ should be exactly the same as the original image $x$. In other words, the effects of $a_t$ and $a_s$ should compensate for each other, which can be regularized by the following cycle consistent loss~\cite{DBLP:journals/corr/ZhuPIE17}.
\begin{equation}\tag{$11$}\label{eq:11}
\mathcal{L}_{cyc}^G=\mathbb{E}_{x,a_t,a_s}[\left\|x-G(G(x,a_t),a_s)\right\|_1]
\end{equation}
where the L1 norm is applied to calculate the difference between $x$ and $x_c$.\vspace{2mm}

\noindent\textbf{Identity Reconstruction Loss}. Another intuition for multi-domain image-to-image translation is that the generated image should also be exactly the same as the input image if $c_t$ equals $c_s$, namely modifying nothing if the target domains are the same as the source domains. In this case, the generated image is denoted by $x_i=G(x,c_s)$, or $x_i=G(x,a_0)$ after considering the action vector, where $a_0$ is the zero action vector indicating that no translation should be conducted. We utilize the L1 norm again and impose the following identity reconstruction loss between $x$ and $x_i$.
\begin{equation}\tag{$12$}\label{eq:12}
\mathcal{L}_{id}^G=\mathbb{E}_{x}[\left\|x-G(x,a_0)\right\|_1]
\end{equation}

\begin{figure*}[!ht]
\centering
\includegraphics[width=0.9\textwidth]{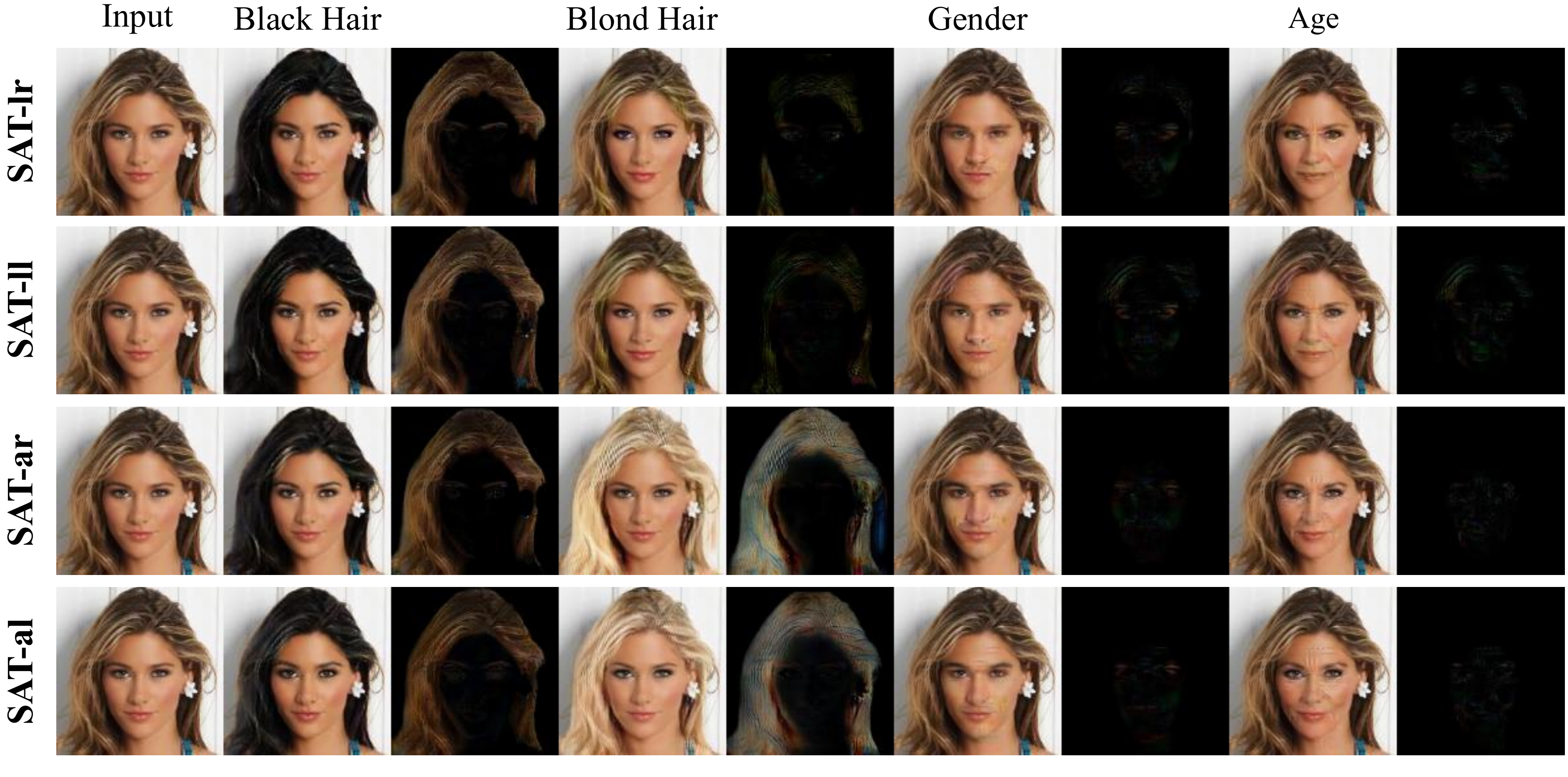}
\caption{Translation results and residual images of the four variants of SAT on the test set. A residual image with fewer non-zero values means better performances of attention. The action vector can better preserve the original identity and generate images that are more related to the target domains.}\label{fourcases}
\end{figure*}

\noindent\textbf{Gradient Penalty}. To stabilize the training process and generate images of higher qualities, we replace Eq.(\ref{eq:7}) and (\ref{eq:8}) with Wasserstein GAN objectives~\cite{DBLP:conf/nips/GulrajaniAADC17}.
\begin{align*}
&\mathcal{L}_{adv}^D=-\mathbb{E}_x[D_{adv}(x)]+\mathbb{E}_{x,a_t}[D_{adv}(G(x,a_t))]\tag{$13$}\label{eq:13}\\
&\mathcal{L}_{adv}^G=-\mathbb{E}_{x,a_t}[D_{adv}(G(x,a_t))]\tag{$14$}\label{eq:14}
\end{align*}

What is more, a gradient penalty loss is imposed on $D$ to enforce the 1-Lipschitz constraint~\cite{DBLP:conf/nips/GulrajaniAADC17}.
\begin{equation}\tag{$15$}\label{eq:15}
\mathcal{L}_{gp}^D=\mathbb{E}_{\bar{x}}[(\left\|\nabla_{\bar{x}}D_{adv}(\bar{x})\right\|_2-1)^2]
\end{equation}
where $\bar{x}$ are uniformly sampled along straight lines between pairs of real and translated samples.
\begin{equation}\tag{$16$}\label{eq:16}
\bar{x}=\epsilon x+(1-\epsilon)G(x,a_t),\epsilon\sim \mathcal{U}(0,1)
\end{equation}

\noindent\textbf{Total Loss}. We combine the losses discussed above to define the total loss functions of $D$ and $G$ to minimize.
\begin{align*}
&\mathcal{L}_D=\mathcal{L}_{adv}^D+\lambda_{cls}\mathcal{L}_{cls}^D+\lambda_{gp}\mathcal{L}_{gp}^D\tag{$17$}\label{eq:17}\\
&\mathcal{L}_G=\mathcal{L}_{adv}^G+\lambda_{cls}\mathcal{L}_{cls}^G+\lambda_{cyc}\mathcal{L}_{cyc}^G+\lambda_{id}\mathcal{L}_{id}^G\tag{$18$}\label{eq:18}
\end{align*}
where $\lambda_{cls},\lambda_{gp},\lambda_{cyc},\lambda_{id}$ are the hyper-parameters to control the weights of different loss terms.

\section{Implementation}

We implement SAT with TensorFlow\footnote{https://www.tensorflow.org/} and conduct all the experiments on a single NVIDIA Tesla P100 GPU.

The network architecture of SAT is thoroughly depicted in Fig.\ref{architecture2}, where blocks of different colors denote different types of neural layers. For the convolution and transposed convolution layers, the attached texts contain the parameters of the convolution kernels, such as $k7n64s1$ denoting a kernel with the kernel size of $7\times 7$, $64$ filters and the stride size of $1$. We apply instance normalization~\cite{DBLP:journals/corr/UlyanovVL16} for $G$ but no normalization for $D$, and the default nonlinearities for $G,D$ are \textit{relu} and \textit{leaky relu} respectively.

The two sub-modules of $G$, TN and AN, both accepts $x,a_t$ and own a \textbf{backbone} respectively, which share the same architecture but are assigned with different parameters. The details of the backbone are also illustrated in Fig.\ref{architecture2}, where (a) (b) denote the raw or latent strategies of combining $x$ or $\tilde{x}$ with $a_t$ by depth-wise concatenation. In the encoder of the backbone, $x$ is downsampled to $\tilde{x}$ by two convolution layers with the stride size of $2$. Six residual blocks are employed and the decoder consists of two transposed convolution layers with the stride size of $2$ for upsampling. 

In order to produce a normalized RGB image in $[-1,1]$, a convolution layer with $3$ filters followed by $tanh$ nonlinearity is appended to TN. In contrast, the backbone of AN is followed by a convolution layer with only $1$ filter and $sigmoid$ nonlinearity to generate an attention mask in $[0,1]$. The outputs of TN and AN are further blended with the input image to obtain the refined result $y$ as Eq.(\ref{eq:6}) defines.

\begin{figure}[!ht]
\centering
\includegraphics[width=0.45\textwidth]{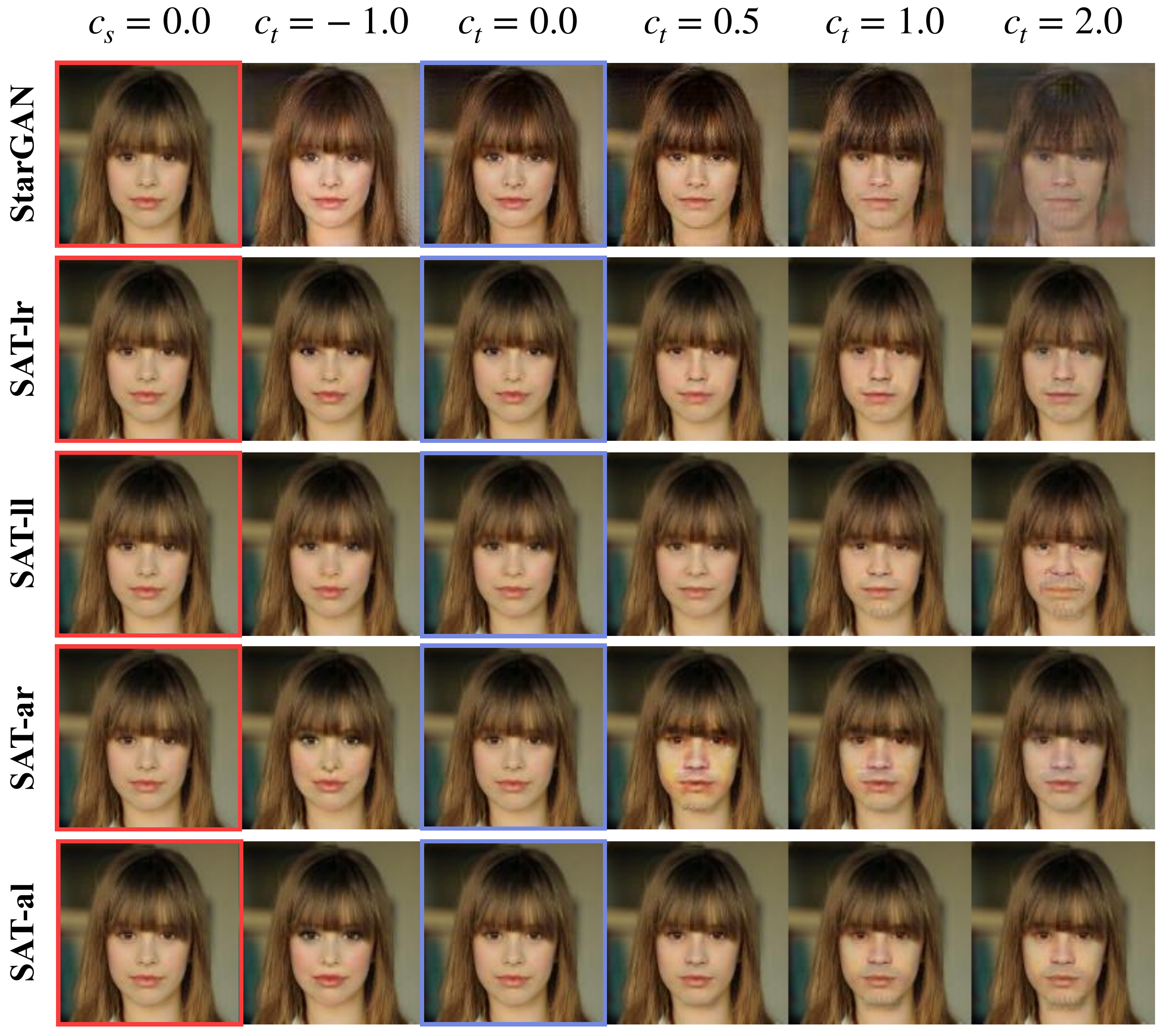}
\caption{Translation results with real-valued vectors on the domain \textit{Male}. The red boxes denote the input images and the blue boxes denote the identity reconstructed images.}\label{realvalues}
\end{figure} 

The network structure of $D$ is relatively simpler, six convolution layers with the stride size of $2$ for downsampling, followed by another two convolution layers for discrimination and classification respectively.

\section{Experiments}

In this section, we perform extensive experiments to demonstrate the superiority of SAT. We first investigate the influences of different settings on the translation results, label or action, raw or latent. Then we explore the feasibility of real-valued vectors and conduct both qualitative as well as quantitative evaluations to compare SAT with existing literature. Lastly, we train SAT again to translate images of higher resolution and visualize the residual images of different domains for better explainability.

\subsection{Datasets}

We utilize the CelebFaces Attributes dataset~\cite{liu2015faceattributes} to conduct experiments for SAT. CelebA contains $202,599$ face images of the size $178\times 218$ from $10,177$ celebrities, each annotated with $40$ binary labels indicating facial attributes like hair color, gender and age. $2,000$ images are randomly selected as the test set and all the other images are used for training. We construct seven domains with the following attributes, hair color~(\textit{black}, \textit{blond}, \textit{brown}), gender~(\textit{male}, \textit{female}) and age~(\textit{young}, \textit{old}).

\subsection{Training}

The images of CelebA are horizontally flipped with a probability of $0.5$, cropped centrally and resized to the resolution of $128\times 128$ for preprocessing. All the parameters of the neural layers are initialized with the Xavier initializer~\cite{DBLP:journals/jmlr/GlorotB10} and we set $\lambda_{cls}=10$,$\lambda_{gp}=10$,$\lambda_{cyc}=10,\lambda_{id}=10$ for all experiments. The Adam~\cite{DBLP:journals/corr/KingmaB14} optimizer is utilized with $\beta_1=0.5,\beta_2=0.999$ and we train SAT on CelebA for $20$ epochs, fixing the learning rate as $0.0001$ for the first $10$ epochs and linearly decaying it to $0$ over the next $10$ epochs.

We set the batch size to $16$ and perform one generator update for every five discriminator updates~\cite{DBLP:conf/nips/GulrajaniAADC17}. For each iteration, we obtain a batch of $x$ and $c_s$ from the training data, randomly shuffle $c_s$ to obtain $c_t$ and calculate $a_t$ accordingly, which imposes $G$ to translate images into various target domains.

\subsection{Different Settings}

We construct four variants of SAT to investigate the influences of different settings.

\begin{itemize}
    \item \textbf{SAT-lr}: the label vector in the raw phase.\vspace{-1mm}
    \item \textbf{SAT-ll}: the label vector in the latent phase.\vspace{-1mm}
    \item \textbf{SAT-ar}: the action vector in the raw phase.\vspace{-1mm}
    \item \textbf{SAT-al}: the action vector in the latent phase.
\end{itemize}

We optimize the above four models on the training images of CelebA and compare their performances on the unseen test set. For each test image, three different operations can be conducted. 1)~\textbf{H}: changing the hair color. 2)~\textbf{G}: inverting the gender~(from \textit{male} to \textit{female} or reversely). 3)~\textbf{A}: inverting the age~(from \textit{young} to \textit{old} or reversely). 

The translation results and the residual images are illustrated in Fig.\ref{fourcases}, where the \textbf{residual image} $\Delta x$ is defined as the difference between the input image $x$ and the translated image $y$, so a residual image with fewer non-zero values means better performances of attention.
\begin{equation}\tag{$19$}\label{eq:19}
\Delta x=\left\|x-y\right\|_1
\end{equation}

As Fig.\ref{fourcases} shows, all the four variants of SAT can produce favorable translation results and the non-zero regions of the residual images intuitively demonstrate the effectiveness of the visual attention. However, we find that the action vector can better preserve the original identity~(see the hair textures in \textit{Black Hair} and the eyes in \textit{Gender}) and generate images more related to the target domains~(see the hair colors in \textit{Blond Hair} and the wrinkles in \textit{Age}). We will further compare the four variants in subsequent sections.

\begin{figure*}[!ht]
\centering
\includegraphics[width=0.86\textwidth]{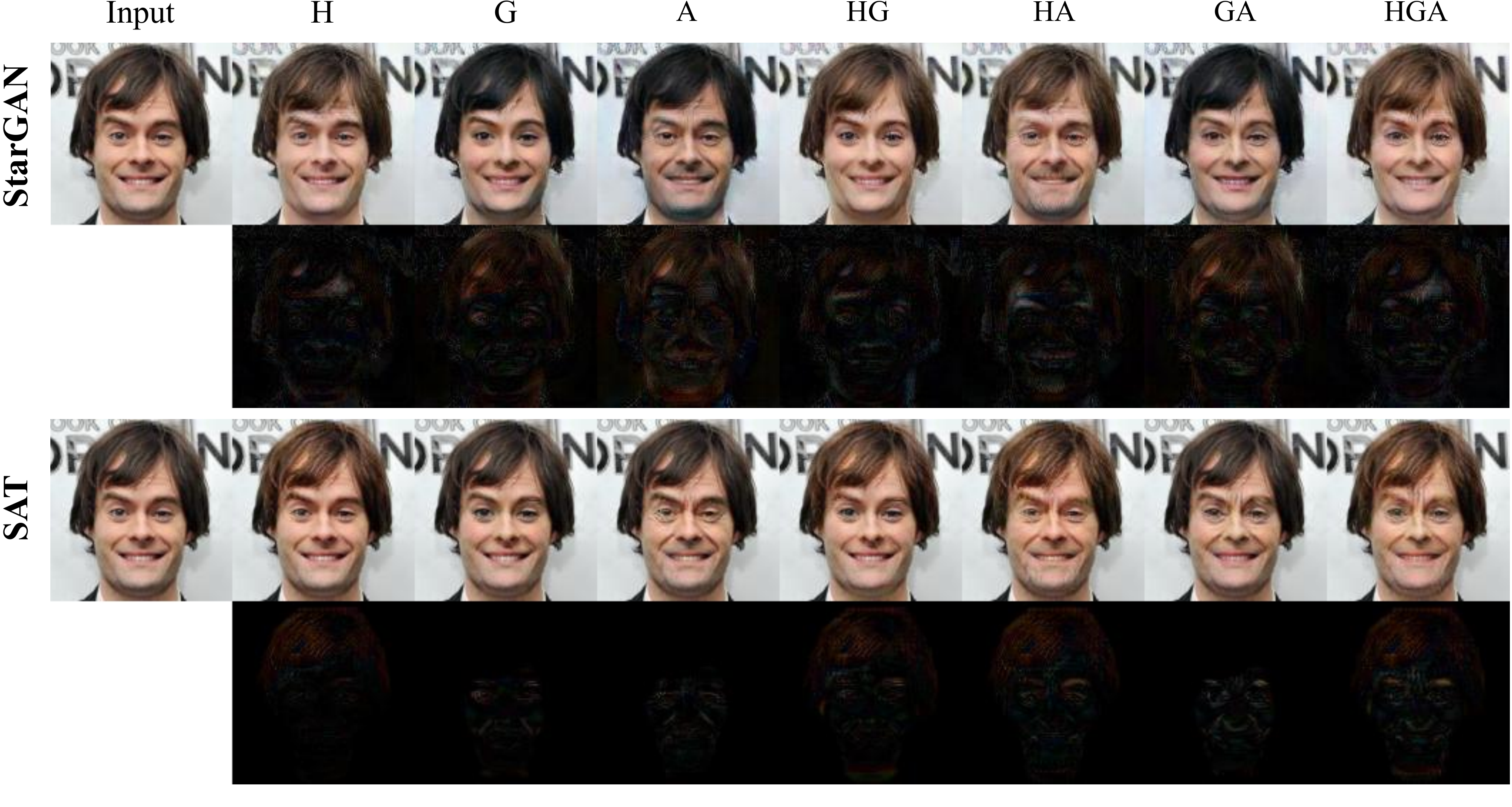}
\caption{Comparisons between StarGAN and SAT. The even rows illustrate the residual images of the two models for each translation task. SAT surpasses StarGAN significantly in both the translation quality of target domains and the preservation of domain-unrelated contents. \textbf{H}: Hair color, \textbf{G}: Gender, \textbf{A}: Age.}\label{compare}
\end{figure*}

\subsection{Baseline Models}

We mainly compare SAT against StarGAN~\cite{DBLP:journals/corr/abs-1711-09020}, which is also devoted to unpaired multi-domain image-to-image translation. The performances of some existing literature on image-to-image translation for two domains like DIAT~\cite{DBLP:journals/corr/LiZZ16e} and CycleGAN~\cite{DBLP:journals/corr/ZhuPIE17} or on facial attribute transfer like IcGAN~\cite{DBLP:journals/corr/PerarnauWRA16} have been detailedly discussed in \cite{DBLP:journals/corr/abs-1711-09020} and surpassed by StarGAN with significant margins, so we ignore them to save space in this paper.

StarGAN utilizes the label vector $c_t$ to convey target domain information and the cycle consistent loss to preserve domain-unrelated contents. $c_t$ is introduced in the raw phase and combined with $x$ by depth-wise concatenation.

\begin{table*}[!ht]
\centering
\begin{tabular}{c c c c c c c c} \hline
\textbf{Operation} & \textbf{H} & \textbf{G} & \textbf{A} & \textbf{HG} & \textbf{HA} & \textbf{GA} & \textbf{HGA} \\ \hline
StarGAN & 18.6\% & 16.7\% & 7.3\% & 23.2\% & 28.4\% & 15.2\% & 32.3\% \\
SAT & \textbf{73.6\%} & \textbf{81.6\%} & \textbf{88.0\%} & \textbf{68.2\%} & \textbf{62.3\%} & \textbf{81.1\%} & \textbf{52.8\%} \\
\textit{Pass} & 7.8\% & 1.7\% & 4.7\% & 8.6\% & 9.3\% & 3.7\% & 14.9\% \\ \hline
\end{tabular}
\caption{Quantitative comparisons of StarGAN and SAT. Each column sums up to 100\%.}\label{tab:1}
\end{table*}

\subsection{Real-Valued Vectors}

We investigate the scalability of StarGAN and the four variants of SAT on real-valued vectors. For example, the domain \textit{Male} can be set as $0$ or $1$ in the label vector, but a robust model should be able to handle abnormal values as well. We test several values of $c_t$ and the results are illustrated in Fig.\ref{realvalues}, where $c_s=0.0$ and $c_t$ can be normal values~($0.0,1.0$) or abnormal values~($-1.0,0.5,2.0$).

For $c_t=0.5$, we observe severe artifacts in the fourth row, indicating the incapability of SAT-ar to handle real-valued vectors. StarGAN fails to translate robustly for extreme values, where the translated images are either over-saturated for $c_t=-1.0$ or under-saturated for $c_t=2.0$, so we conjecture that StarGAN wrongly correlates the domain \textit{Male} with the saturation of images. The domain-unrelated regions like the background are also influenced and the image is even badly blurred when $c_t=2.0$.

In contrast, the domain-unrelated regions are well preserved in all the four variants of SAT, which are mainly owing to the effective visual attention. Constrained by the identity reconstruction loss, the four variants can also perfectly preserve the original identity when $c_t=c_s$~(see the red boxes and the blue boxes in Fig.\ref{realvalues}). However, SAT-ll fails to perform robustly and wrongly translates the girl into an old man for $c_t=2.0$, indicating that SAT-ll confuses the semantics of the domain \textit{Male} with the domain \textit{Old}. Both SAT-lr and SAT-al can always achieve reasonable results for all values of $c_t$, but we are delighted to discover that SAT-al is even capable of naturally adding some auxiliary features when dealing with extreme values, such as the makeups for $c_t=-1.0$ and the beards for $c_t=2.0$. The above differences are also observed on other test images.

As a result, we arrive at the following three conclusions. 1) It is better to introduce the \textit{label} vector in the \textit{raw} phase. 2) It is better to introduce the \textit{action} vector in the \textit{latent} phase. 3) SAT-al is superior to the other three variants. We will use SAT-al as the default choice of SAT for subsequent experiments.

\begin{figure*}[!ht]
\centering
\includegraphics[width=0.89\textwidth]{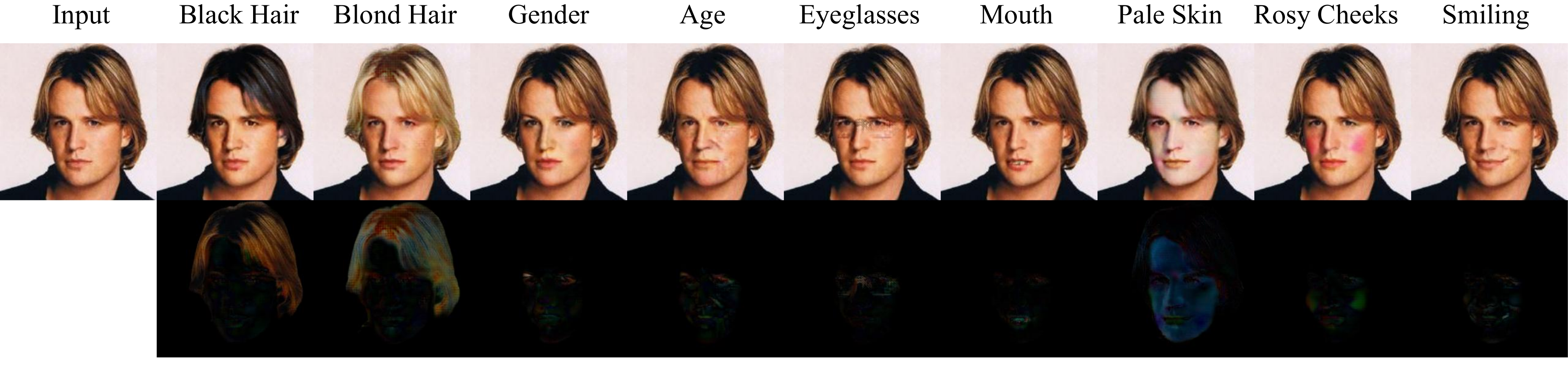}
\caption{Translation results and residual images of the resolution $256\times 256$ for more domains.}\label{higher}
\end{figure*}

\begin{figure*}[!ht]
\centering
\includegraphics[width=0.9\textwidth]{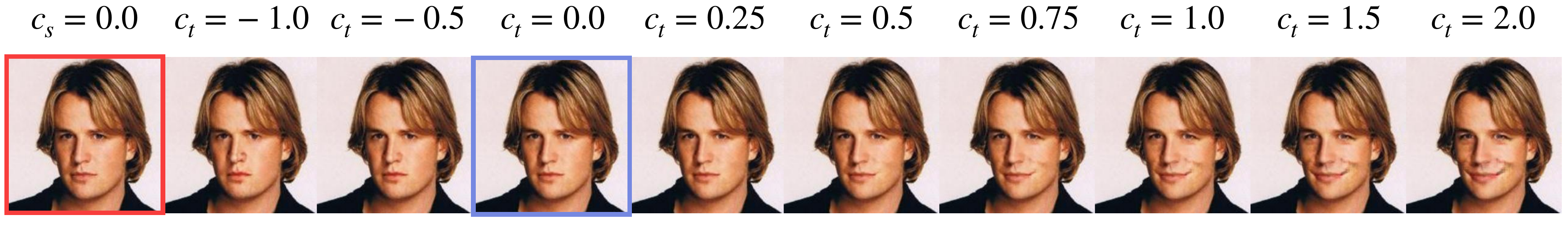}
\caption{Translation results with real-valued vectors on the domain \textit{Smiling}.}\label{smile}
\end{figure*}

\subsection{Qualitative Evaluation}

Based on H, G, A, we further construct another four operations for each test image, HG, HA, GA, HGA, to cover translations for multiple domains. Fig.\ref{compare} illustrates the qualitative results of StarGAN and SAT for different operations, where the first column shows the input image, followed by three columns for single-domain translation and four columns for multi-domain translation.

As Fig.\ref{compare} shows, StarGAN can generate visually satisfactory results, but unavoidably influences the domain-unrelated contents. For example, the background is changed for almost all operations according to the residual images, and the hair color wrongly turns black when the translation tasks are G, A and GA. StarGAN fails to disentangle the patterns of different domains and produces undesired changes, which may degrade the translation performances for certain operations.

The above problems of StarGAN are not observed in SAT. The residual images demonstrate the capability of SAT to focus only on the domain-related regions, such as the hair when translating hair color and the face when translating gender or age. Due to the effective action vector, SAT can capture the semantics of each domain and generate realistic images with reasonable details.

\subsection{Quantitative Evaluation}

For each of the 2,000 images in the test set, we perform the above seven operations and obtain a pair of candidates from the two models. The quantitative evaluation is conducted in a crowd-sourcing manner, where the volunteers are instructed to select the better one based on three criterias, the perceptual realism, the quality of translation for target domains, and the preservation of original identity.

The statistics are reported in Table \ref{tab:1}, where \textit{Pass} means the volunteers cannot make a choice when the two images are as good or as bad. The translation results depend heavily on the quality of the input images, and both models may perform poorly when the input images are blurred and under-saturated, which accounts for a large proportion of \textit{Pass}. For single-domain translation, SAT surpasses StarGAN by winning $66.9\%$ more votes on average. For the more complicated tasks of multi-domain translation, the percentages of \textit{Pass} increase a lot, but SAT is still superior to StarGAN with a significant margin of $41.3\%$ on average.

\subsection{Higher Resolution}

We process the images of CelebA to the size $256\times 256$ and train SAT again to translate images of higher resolution. We construct more domains based on the following ten attributes, \textit{black hair, blond hair, brown hair, male, young, eyeglasses, mouth slightly open, pale skin, rosy cheeks, smiling}, to investigate the robustness of our model.

The translation results and the residual images are shown in Fig.\ref{higher}, which prove the effectiveness of SAT to correctly translate images for the target domains by only changing the related contents. For the domain \textit{Smiling}, we inspect SAT with various values of $c_t$ and illustrate the results in Fig.\ref{smile}, which demonstrates that SAT can also be utilized to achieve other tasks like facial expression synthesis~\cite{DBLP:conf/eccv/PumarolaAMSM18}.

\section{Conclusion}

In this paper, we propose SAT, an unified and explainable generative adversarial network for unpaired multi-domain image-to-image translation with a single generator. Based on the label vector, we propose the action vector to convey target domain information and introduce it in the latent phase. The visual attention is utilized so that only the domain-related regions are translated with the other preserved. Extensive experiments demonstrate the superiority of our model and the residual images better explain what SAT attends when translating images for different domains.

{\small
\bibliographystyle{ieee}
\bibliography{egbib}

\begin{thebibliography}{1}\itemsep=-1pt

\bibitem{doi:10.1080/02699930903485076}
O.~Langner, R.~Dotsch, G.~Bijlstra, D.~H.~J. Wigboldus, S.~T. Hawk, and A.~van
  Knippenberg.
\newblock Presentation and validation of the radboud faces database.
\newblock {\em Cognition and Emotion}, 24(8):1377--1388, 2010.

\bibitem{DBLP:journals/corr/ZhuPIE17}
J.~Zhu, T.~Park, P.~Isola, and A.~A. Efros.
\newblock Unpaired image-to-image translation using cycle-consistent
  adversarial networks.
\newblock {\em CoRR}, abs/1703.10593, 2017.

\end{thebibliography}


\begin{thebibliography}{10}\itemsep=-1pt

\bibitem{DBLP:journals/corr/abs-1711-09020}
Y.~Choi, M.~Choi, M.~Kim, J.~Ha, S.~Kim, and J.~Choo.
\newblock Stargan: Unified generative adversarial networks for multi-domain
  image-to-image translation.
\newblock {\em CoRR}, abs/1711.09020, 2017.

\bibitem{DBLP:journals/jmlr/GlorotB10}
X.~Glorot and Y.~Bengio.
\newblock Understanding the difficulty of training deep feedforward neural
  networks.
\newblock In {\em {AISTATS}}, pages 249--256, 2010.

\bibitem{DBLP:conf/nips/GoodfellowPMXWOCB14}
I.~J. Goodfellow, J.~Pouget{-}Abadie, M.~Mirza, B.~Xu, D.~Warde{-}Farley,
  S.~Ozair, A.~C. Courville, and Y.~Bengio.
\newblock Generative adversarial nets.
\newblock In {\em {NIPS}}, pages 2672--2680, 2014.

\bibitem{DBLP:conf/nips/GulrajaniAADC17}
I.~Gulrajani, F.~Ahmed, M.~Arjovsky, V.~Dumoulin, and A.~C. Courville.
\newblock Improved training of wasserstein gans.
\newblock In {\em {NIPS}}, pages 5769--5779, 2017.

\bibitem{DBLP:conf/cvpr/HeZRS16}
K.~He, X.~Zhang, S.~Ren, and J.~Sun.
\newblock Deep residual learning for image recognition.
\newblock In {\em {CVPR}}, pages 770--778, 2016.

\bibitem{DBLP:conf/cvpr/IsolaZZE17}
P.~Isola, J.~Zhu, T.~Zhou, and A.~A. Efros.
\newblock Image-to-image translation with conditional adversarial networks.
\newblock In {\em {CVPR}}, pages 5967--5976, 2017.

\bibitem{DBLP:conf/icml/KimCKLK17}
T.~Kim, M.~Cha, H.~Kim, J.~K. Lee, and J.~Kim.
\newblock Learning to discover cross-domain relations with generative
  adversarial networks.
\newblock In {\em {ICML}}, pages 1857--1865, 2017.

\bibitem{DBLP:journals/corr/KingmaB14}
D.~P. Kingma and J.~Ba.
\newblock Adam: {A} method for stochastic optimization.
\newblock {\em CoRR}, abs/1412.6980, 2014.

\bibitem{DBLP:conf/cvpr/LedigTHCCAATTWS17}
C.~Ledig, L.~Theis, F.~Huszar, J.~Caballero, A.~Cunningham, A.~Acosta, A.~P.
  Aitken, A.~Tejani, J.~Totz, Z.~Wang, and W.~Shi.
\newblock Photo-realistic single image super-resolution using a generative
  adversarial network.
\newblock In {\em {CVPR}}, pages 105--114, 2017.

\bibitem{DBLP:journals/corr/LiZZ16e}
M.~Li, W.~Zuo, and D.~Zhang.
\newblock Deep identity-aware transfer of facial attributes.
\newblock {\em CoRR}, abs/1610.05586, 2016.

\bibitem{DBLP:conf/nips/LiuBK17}
M.~Liu, T.~Breuel, and J.~Kautz.
\newblock Unsupervised image-to-image translation networks.
\newblock In {\em {NIPS}}, pages 700--708, 2017.

\bibitem{liu2015faceattributes}
Z.~Liu, P.~Luo, X.~Wang, and X.~Tang.
\newblock Deep learning face attributes in the wild.
\newblock In {\em {ICCV}}, 2015.

\bibitem{DBLP:journals/corr/MirzaO14}
M.~Mirza and S.~Osindero.
\newblock Conditional generative adversarial nets.
\newblock {\em CoRR}, abs/1411.1784, 2014.

\bibitem{DBLP:conf/icml/OdenaOS17}
A.~Odena, C.~Olah, and J.~Shlens.
\newblock Conditional image synthesis with auxiliary classifier gans.
\newblock In {\em {ICML}}, pages 2642--2651, 2017.

\bibitem{DBLP:journals/corr/PerarnauWRA16}
G.~Perarnau, J.~van~de Weijer, B.~Raducanu, and J.~M. {\'{A}}lvarez.
\newblock Invertible conditional gans for image editing.
\newblock {\em CoRR}, abs/1611.06355, 2016.

\bibitem{DBLP:conf/eccv/PumarolaAMSM18}
A.~Pumarola, A.~Agudo, A.~M. Martinez, A.~Sanfeliu, and F.~Moreno{-}Noguer.
\newblock Ganimation: Anatomically-aware facial animation from a single image.
\newblock In {\em {ECCV}}, pages 835--851, 2018.

\bibitem{DBLP:journals/corr/RadfordMC15}
A.~Radford, L.~Metz, and S.~Chintala.
\newblock Unsupervised representation learning with deep convolutional
  generative adversarial networks.
\newblock {\em CoRR}, abs/1511.06434, 2015.

\bibitem{DBLP:journals/corr/UlyanovVL16}
D.~Ulyanov, A.~Vedaldi, and V.~S. Lempitsky.
\newblock Instance normalization: The missing ingredient for fast stylization.
\newblock {\em CoRR}, abs/1607.08022, 2016.

\bibitem{DBLP:conf/nips/VaswaniSPUJGKP17}
A.~Vaswani, N.~Shazeer, N.~Parmar, J.~Uszkoreit, L.~Jones, A.~N. Gomez,
  L.~Kaiser, and I.~Polosukhin.
\newblock Attention is all you need.
\newblock In {\em {NIPS}}, pages 6000--6010, 2017.

\bibitem{DBLP:conf/icml/XuBKCCSZB15}
K.~Xu, J.~Ba, R.~Kiros, K.~Cho, A.~C. Courville, R.~Salakhutdinov, R.~S. Zemel,
  and Y.~Bengio.
\newblock Show, attend and tell: Neural image caption generation with visual
  attention.
\newblock In {\em {ICML}}, pages 2048--2057, 2015.

\bibitem{DBLP:journals/corr/abs-1711-10485}
T.~Xu, P.~Zhang, Q.~Huang, H.~Zhang, Z.~Gan, X.~Huang, and X.~He.
\newblock Attngan: Fine-grained text to image generation with attentional
  generative adversarial networks.
\newblock {\em CoRR}, abs/1711.10485, 2017.

\bibitem{DBLP:conf/eccv/ZhangKSC18}
G.~Zhang, M.~Kan, S.~Shan, and X.~Chen.
\newblock Generative adversarial network with spatial attention for face
  attribute editing.
\newblock In {\em {ECCV}}, pages 422--437, 2018.

\bibitem{DBLP:journals/corr/abs-1805-08318}
H.~Zhang, I.~J. Goodfellow, D.~N. Metaxas, and A.~Odena.
\newblock Self-attention generative adversarial networks.
\newblock {\em CoRR}, abs/1805.08318, 2018.

\bibitem{DBLP:conf/iccv/ZhangXL17}
H.~Zhang, T.~Xu, and H.~Li.
\newblock Stackgan: Text to photo-realistic image synthesis with stacked
  generative adversarial networks.
\newblock In {\em {ICCV}}, pages 5908--5916, 2017.

\bibitem{DBLP:journals/corr/abs-1710-10916}
H.~Zhang, T.~Xu, H.~Li, S.~Zhang, X.~Wang, X.~Huang, and D.~N. Metaxas.
\newblock Stackgan++: Realistic image synthesis with stacked generative
  adversarial networks.
\newblock {\em CoRR}, abs/1710.10916, 2017.

\bibitem{DBLP:journals/corr/ZhuPIE17}
J.~Zhu, T.~Park, P.~Isola, and A.~A. Efros.
\newblock Unpaired image-to-image translation using cycle-consistent
  adversarial networks.
\newblock {\em CoRR}, abs/1703.10593, 2017.

\end{thebibliography}
}

\end{document}